\documentclass[11pt]{article}
\usepackage{eamt17}
\usepackage{times}
\usepackage{url}
\usepackage{latexsym}

\title{OpenNMT: Open-source Toolkit for Neural Machine Translation}

\author{Guillaume Klein$^\dagger$, Yoon Kim$^*$, Yuntian Deng$^*$, Josep Crego$^\dagger$\\{\bf Jean Senellart}$^\dagger$, {\bf Alexander M. Rush}$^*$ \\ Harvard University$^*$, SYSTRAN $^\dagger$}

\date{}

\begin{document}
\maketitle
\begin{abstract}
  We introduce an open-source toolkit for neural machine translation
  (NMT) to support research into model architectures, feature representations,
  and source modalities, while maintaining competitive performance, modularity
  and reasonable training requirements.
\end{abstract}

\section{Introduction}

Neural machine translation has become a set of standardised approaches that has led to remarkable improvements, particularly in terms of human evaluation. It has now been successfully applied in production environment by major translation technology providers.

\textit{OpenNMT}\footnote{\url{http://opennmt.net}} is an open (MIT licensed) and joint initiative by SYSTRAN and the Harvard NLP group to develop a NMT toolkit for researchers and engineers to benchmark against, learn from, extend and build upon. It focuses on providing a production-grade system with an extensive set of model and training options to cover a large set of needs of academia and industry.

\section{Description}

\textit{OpenNMT} implements the complete sequence-to-sequence approach that achieved state-of-the-art results in many tasks including machine translation. Based on the Torch framework, this model comes with many extensions that are known useful including multi-layer RNN, attention, bidirectional encoder, word features, input feeding, residual connections, beam search, and several others. 
The toolkit also provides various options to customize the training process depending on the task and data with multi-GPU support, re-training, data sampling and learning rate decay strategies.

Toolkits like \textit{Nematus}\footnote{\url{https://github.com/rsennrich/nematus}} or Google's \textit{seq2seq}\footnote{\url{https://github.com/google/seq2seq}} share similar goals and implementation but with frequent limitations on efficiency, tooling, features or documentation which \textit{OpenNMT} tries to solve.

\section{Ecosystem}

More than the core project, \textit{OpenNMT} aims to propose an ecosystem around NMT and sequence modelling. It comes with an optimised C++ inference engine based on the Eigen library to make deployment and integration of models easy and efficient. The library has also been used on multiple tasks, including image-to-text, speech-to-text and summarisation. We also provide recipes to automatise the training process, demo servers to quickly showcase results and a benchmark platform\footnote{\url{http://nmt-benchmark.net/}} to compare approaches.

\section{Community}

\textit{OpenNMT} is also a community\footnote{\url{http://forum.opennmt.net/}} providing various supports on using the project, addressing specific training processes and discussing the current and future state of neural machine translation research and development. The online forum counts more than 100 users and the project has been starred by over 1,000 users on GitHub.

\section{Conclusion}

We introduce \textit{OpenNMT}, a research toolkit for neural MT that prioritises efficiency and modularity. We hope to maintain strong machine translation results at the research frontier, providing a stable framework for production use while enlarging an active and motivated community.


\end{document}